\documentclass{article} 
\usepackage{iclr2025_conference,times}

\usepackage{microtype}
\usepackage{graphicx}
\usepackage{subfigure}
\usepackage{booktabs} 

\usepackage{amsfonts}
\usepackage{nicefrac}
\usepackage{xcolor}
\usepackage{algorithm}
\usepackage{algorithmic}
\usepackage{subcaption}
\usepackage{ifsym}
\usepackage{diagbox}
\usepackage{float}
\usepackage{multirow}
\usepackage{epsfig}
\usepackage{marvosym}
\usepackage{amsmath}
\usepackage{amssymb}
\usepackage{mathtools}
\usepackage{amsthm}
\usepackage{wasysym}
\usepackage{enumitem}
\usepackage{tcolorbox}
\usepackage{makecell}
\usepackage{colortbl}
\usepackage{tikz}
\usepackage{bbm}
\usepackage{tabularx}

\definecolor{baselinecolor}{gray}{.9}
\definecolor{myred}{rgb}{0.992,0.9576,0.932}
\definecolor{mydred}{rgb}{0.992,0.915,0.892}
\definecolor{mypink}{rgb}{1,0.95,0.962}
\definecolor{myyellow}{rgb}{0.99,1,0.78}
\definecolor{myredd}{rgb}{0.992,0.9076,0.63}
\definecolor{mydredd}{rgb}{0.96,0.72,0.72}
\definecolor{mypinkd}{rgb}{0.98,0.75,0.952}

\definecolor{myg4}{rgb}{0.98,0.98,0.98}
\definecolor{myg5}{rgb}{0.96,0.96,0.96}
\definecolor{myg6}{rgb}{0.94,0.94,0.94}
\definecolor{myg7}{rgb}{0.92,0.92,0.92}
\definecolor{myg8}{rgb}{0.9,0.9,0.9}
\definecolor{myg9}{rgb}{0.88,0.88,0.88}
\definecolor{myg10}{rgb}{0.86,0.86,0.86}
\definecolor{myg11}{rgb}{0.84,0.84,0.84}
\definecolor{myg12}{rgb}{0.82,0.82,0.82}

\usepackage{amsmath,amsfonts,bm}

\def\eqref#1{equation~\ref{#1}}

\def\1{\bm{1}}

\DeclareMathAlphabet{\mathsfit}{\encodingdefault}{\sfdefault}{m}{sl}
\SetMathAlphabet{\mathsfit}{bold}{\encodingdefault}{\sfdefault}{bx}{n}

\usepackage{hyperref}
\usepackage{url}

\title{Model Assembly Learning with Heterogeneous Layer Weight Merging}

\author{
  \textbf{Yi-Kai Zhang}$^{1,2}$ \ \ \ \textbf{Jin Wang}$^3$ \ \ \ \textbf{Xu-Xiang Zhong}$^{1,2}$ \ \ \ \textbf{De-Chuan Zhan}$^{1,2}$ \ \ \ \textbf{Han-Jia Ye}$^{1,2}$\thanks{Corresponding author, email: yehj@lamda.nju.edu.cn.}\\
  $^1$School of Artificial Intelligence, Nanjing University\\
  $^2$National Key Laboratory for Novel Software Technology, Nanjing University\\ $^3$Yingcai Honors College, University of Electronic Science and Technology of China \\
}

\iclrfinalcopy 
\begin{document}

\maketitle

\begin{abstract}

Model merging acquires general capabilities without extra data or training by combining multiple models' parameters.
Previous approaches achieve linear mode connectivity by aligning parameters into the same loss basin using permutation invariance.
In this paper, we introduce Model Assembly Learning (MAL), a novel paradigm for model merging that iteratively integrates parameters from diverse models in an open-ended model zoo to enhance the base model's capabilities. Unlike previous works that require identical architectures, MAL allows the merging of heterogeneous architectures and selective parameters across layers. Specifically, the base model can incorporate parameters from different layers of multiple pre-trained models.
We systematically investigate the conditions and fundamental settings of heterogeneous parameter merging, addressing all possible mismatches in layer widths between the base and target models. Furthermore, we establish key laws and provide practical guidelines for effectively implementing MAL.
\end{abstract}

\section{Introduction}

Deep learning has witnessed remarkable advancements across various domains, primarily driven by stochastic gradient descent (SGD) optimization that efficiently tackles non-convex optimization challenges. However, as the \textit{pretraining-finetuning} paradigm becomes more popular, extending model capabilities by further training on novel data presents significant challenges. These include catastrophic forgetting~\citep{DBLP:conf/acl/YadavB23,DBLP:journals/pami/ZhouWQYZL24}, poor out-of-distribution generalization~\citep{DBLP:conf/iclr/Jin0P023,DBLP:journals/corr/abs-2403-01874} of the model, with high annotation costs and privacy constraints~\citep{DBLP:conf/aistats/McMahanMRHA17} on target data. While continual learning helps address these issues to some extent, its complexity and high computational cost make real-world deployment challenging~\citep{DBLP:conf/nips/FiftyAZYAF21,DBLP:conf/iclr/Adel0T20}. This raises a crucial question:  
Can we leverage pre-trained parameters from a large-scale model zoo to extract new knowledge without requiring additional data by discovering patterns in different parameter spaces?

Several studies~\citep{DBLP:conf/nips/MatenaR22,DBLP:conf/iclr/Jin0P023,DBLP:journals/corr/abs-2406-11617,DBLP:conf/eccv/JangYH24,DBLP:conf/icml/Yu0Y0L24} have explored merging parameters with the same model architecture. Model Soups~\citep{DBLP:conf/icml/WortsmanIGRLMNF22} directly averages model parameters that share the same pre-trained initialization but are fine-tuned with different hyperparameters. Other approaches relax the assumption of the same initialization and leverage the permutation invariance to align and merge models trained on the same dataset. These methods optimize convex combinations of aligned model parameters, ensuring that along the merging path, performance remains non-decreasing relative to a direct interpolation between the base and target models---a property known as linear mode connectivity (LMC)~\citep{DBLP:conf/icml/FrankleD0C20,DBLP:conf/iclr/EntezariSSN22}.
Expanding beyond this, \cite{DBLP:conf/iclr/StoicaBBRHH24} introduced \textit{ZipIt!}, which selectively merges relevant features and partial parameters, extending applicability to models trained on different datasets but all of the above with the same architecture.

In this paper, we study model merging for heterogeneous parameters and propose Model Assembly Learning (MAL). Given a base model and a large-scale zoo of pre-trained models from different tasks and architectures, the base model freely searches for and integrates independent layer parameters from the repository, guided by prior laws from \textit{a bag of laws}. Each merging step may assemble layers from multiple pre-trained models, iteratively expanding the base capabilities without requiring additional target data.
Specifically, we break down the process into layer-wise parameter merging. Unlike homogeneous model merging, we relax the requirement for entire model output invariance and instead focus on layer output invariance as a weaker but sufficient condition. The LMC is re-defined more flexibly: As layers from the base model are gradually replaced with merged parameters, the model retains its original domain performance up to a critical convex combination threshold.

A major obstacle in merging heterogeneous models is inconsistent layer widths. We introduce a generalized permutation transformation that aligns parameters while preserving key information to overcome this. This process includes: \textbf{1}) Zero-padding and permutation to harmonize layer dimensions, enabling knowledge transfer from models with differing widths; \textbf{2}) Bidirectional alignment between base and target layers, where the base model actively permutes its parameters to align with those of the target model optimally.
Our contributions are:
\begin{itemize}[topsep=0pt,leftmargin=12pt]
\item We present Model Assembly Learning, a novel paradigm that enables flexible, layer-wise parameter merging to integrate diverse trained knowledge.
\item We introduce a heterogeneous merging strategy, addressing layer width mismatches through bidirectional permutation to facilitate effective parameter transfer.
\item We validate our approach through assembly learning on a large-scale model zoo, analyzing heterogeneous layer fusion patterns across 30 architectures spanning 5 categories.
\end{itemize}

\section{Background}

\textbf{Merging of Convex Combinations and Linear Mode Connectivity.} Neural networks with the same architecture naturally have corresponding layers and widths. Suppose they meet certain conditions, such as sharing part of the optimization trajectory~\citep{DBLP:conf/icml/FrankleD0C20,DBLP:conf/nips/NeyshaburSZ20}. In that case, the convex combination of their weights can be merged with any factor without increasing the loss~\citep{DBLP:conf/acml/ONeillSG21,DBLP:journals/corr/abs-2312-16240,DBLP:conf/icml/HoroiCBW24}. Given model $A$ and $B$ with weights $\mathbf{W}^A, \mathbf{W}^B$ and factor $\lambda$, we have:

\textbf{Definition 1} (Loss barrier~\citep{DBLP:conf/icml/FrankleD0C20}).
Given two weights $\mathbf{W}^A$, $\mathbf{W}^B$ such that $\mathcal{L}\left(\mathbf{W}^A\right) \approx \mathcal{L}\left(\mathbf{W}^B\right)$,  the \textit{loss barrier} is defined as
\begin{equation}
    \footnotesize
    \max_{\lambda \in [0,1]} \mathcal{L}\left(\left(1 - \lambda\right) \mathbf{W}^A + \lambda \mathbf{W}^B\right) - \frac{1}{2} \left( \mathcal{L} \left(\Theta_A\right) + L\left(\Theta_B\right) \right)\;.
\end{equation}

When the non-negative loss barrier is sufficiently small, an effective loss landscape path---known as linear mode connectivity (LMC)---is established between the two models. However, when models have different initializations, directly merging them can lead to a catastrophic loss barrier. Some works~\citep{DBLP:conf/nips/SinghJ20,DBLP:conf/iclr/AinsworthHS23,DBLP:conf/iclr/ImfeldGGHAS24} leverage the permutation symmetries of weight space, rearranging neurons while preserving output consistency. This alignment maps corresponding neurons to the same parameter space, effectively restoring LMC.
We define a permutation matrix as $\boldsymbol{P}$, which is generally a square matrix where each row and column sums to 1. For an $L$-layer model, at the $l$-th layer, we have:
\begin{equation}
    \boldsymbol{h}_{l+1}=\boldsymbol{P}^{\top} \boldsymbol{P} \cdot \boldsymbol{h}_{l+1}=\boldsymbol{P}^{\top} \boldsymbol{P} \cdot \sigma\left(\mathbf{W}_{l} \cdot  \boldsymbol{h}_{l}+\boldsymbol{b}_{l}\right)=\boldsymbol{P}^{\top} \sigma\left(\boldsymbol{P} \mathbf{W}_{l} \cdot \boldsymbol{h}_{l}+\boldsymbol{P} \cdot \boldsymbol{b}_{l}\right)\;,
\end{equation}
where $\boldsymbol{h}_{l}$ represents the output of layer $l$.
At this point, $\boldsymbol{P}^{\top}$ combines with the $\mathbf{W}_{l+1}$ of the next layer, and we obtain a functionally equivalent model as:
\begin{equation}
    \mathbb{P} \left(\mathbf{W}_l \right) = \boldsymbol{P}_{l+1} \mathbf{W}_l \boldsymbol{P}_{l} \, , \, \, \, \mathbb{P} \left(\boldsymbol{b}_l\right) = \boldsymbol{P}_{l+1} \boldsymbol{b}_l \,,  \, \, \, \boldsymbol{P}_{1} = \boldsymbol{P}_{n + 1} = \mathbf{I}\;.
\end{equation}
Some parameter merging methods~\citep{DBLP:journals/corr/abs-2403-19390,xu2024trainingfreeheterogeneousmodelmerging,DBLP:conf/emnlp/GoddardSEMKBMS24} determine an appropriate $\boldsymbol{P}$ such that $\mathbf{W}^A$ and the transformed parameters, denoted as $\mathbb{P}(\mathbf{W}^B)$, together establish the LMC condition.

Model merging plays a crucial role in various domains, including federated learning~\citep{DBLP:conf/mlsys/LiSZSTS20}, model compression~\citep{DBLP:journals/corr/abs-2407-09590}, and multimodal continual adaptation~\citep{DBLP:conf/nips/LiSGJXH21}. In this paper, we explore a more generalized scenario, \textit{i.e.}, Model Assembly Learning (MAL), that removes constraints on model initialization and architecture. MAL iteratively integrates multiple models' parameters into the base model, merging them layer by layer to extract valuable knowledge from each. We employ LMC to validate the effectiveness of this paradigm.

\section{Method}

\textbf{Motivation:} The fundamental principle shows that two models performing the same task should learn similar features. Since similar weights produce similar features, we align corresponding weights to facilitate knowledge transfer across layers of different models. By continuously merging proper layer-wise weights from the model zoo, we form the paradigm called Model Assembly Learning (MAL) by continuously merging useful layer-wise weights from the model zoo.

Previous studies~\citep{DBLP:conf/iclr/AinsworthHS23} on merging models with identical architectures have aimed to align parameters at each layer for a globally optimal solution. However, this problem is NP-hard. Therefore, they typically relax the objective to enforce weight similarity at the layer level, optimizing:
\begin{equation}
\underset{\mathbb{P}}{\arg \min} \| \mathbf{W}^{A}_l - \mathbb{P} \left( \mathbf{W}^{B}_l \right) \|^2 = \underset{\boldsymbol{P}}{\arg \max} \left\langle\mathbf{W}_{l}^{A}, \boldsymbol{P}_{l} \mathbf{W}_{l}^{B} \boldsymbol{P}_{l-1}^{\top}\right\rangle_F \;,
\end{equation}
where the bias term is omitted for simplicity. Since each layer can be regarded as a subnetwork with specific feature inputs and outputs, we focus on merging the subsets of the parameters, effectively assembling them into a base model.

\begin{figure}[t]
    \centering
    \includegraphics[width=\textwidth]{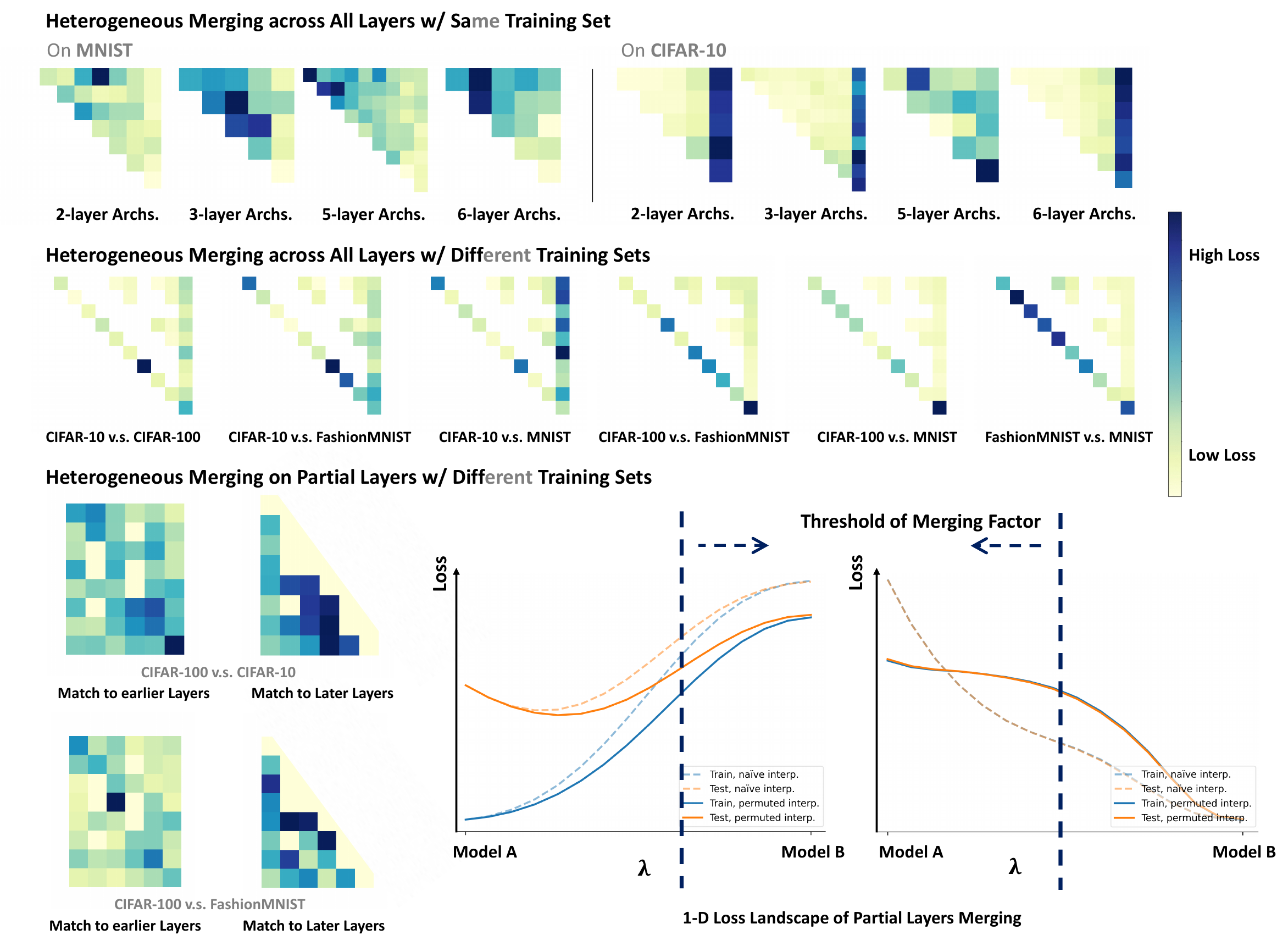}
    \caption{Results with merging to a large-scale model zoo with various architectures and training sets. In each heatmap, the horizontal and vertical axes represent different model architectures, and the cell values indicate the mean LMC loss during merging as $\lambda$ varies (lighter colors denote better merging). The first two rows display results for heterogeneous models on the same/different datasets, while the third and fourth rows show merging results at different positions in \textit{selected layers}. The two plots in the lower right illustrate the loss during the merging process; notably, before the merging factor threshold, merging heterogeneous parameters can maintain a certain LMC (see the left plot).}
    \label{fig:results}
\end{figure}

\textbf{Generalized Permutation for Shape Mismatches.} In practice, layer parameters of the open-end model zoo often differ in shape. We introduce a generalized permutation operation, represented by $\boldsymbol{P} \in \mathbb{R}^{m \times n}$, where $m \neq n$, meaning $\boldsymbol{P}$ is not necessarily a square matrix. The essence of this transformation is to allow structural adjustments by padding the smaller dimension with zeros and align to the larger one.
Let $\mathbf{W}^{A} \in \mathbb{R}^{m_1 \times n_1}$ and $\mathbf{W}^{B} \in \mathbb{R}^{m_2 \times n_2}$, where the dimensions of the parameters in the corresponding layers are not exactly the same. This discrepancy induces a \textit{partial order relation} between the shape of weight matrices, which we classify into two cases:
\begin{itemize}[topsep=0pt,leftmargin=12pt]
\item \textbf{1}) Size-compatible case as $(m_1 - m_2) \times (n_1 - n_2) \geq 0$: If one weight matrix strictly contains the dimensions of the other, the layers are size-comparable. In this case, we apply the generalized permutation $\boldsymbol{P}$ on the smaller matrix (denoted as $\mathbf{W}^B$), adjusting the row/column dimension accordingly. The optimization objective remains as:
\begin{equation}
    \underset{\boldsymbol{P}_{l}}{\arg \max }\left\langle\mathbf{W}_{l}^{A}, \, \boldsymbol{P}_{l} \mathbf{W}_{l}^{B} \boldsymbol{P}_{l-1}^{\top}\right\rangle_F+\left\langle\mathbf{W}_{l+1}^{A}, \, \boldsymbol{P}_{l+1} \mathbf{W}_{l+1}^{B} \boldsymbol{P}_{l}^{\top}\right\rangle_F\;.
\end{equation}

\item \textbf{2}) Size-incompatible case as $(m_1 - m_2) \times (n_1 - n_2) < 0$: If the layers' dimensions do not follow a strict order, a direct transformation is not possible. Instead, we introduce bidirectional permutation operations: $\boldsymbol{P}$ on $\mathbf{W}^{B}$ and $\boldsymbol{Q}$ on $\mathbf{W}^{A}$.
The corresponding optimization objective is:
\begin{equation}
\label{eq:incompatible}
\begin{aligned}
     & \underset{\boldsymbol{P}_l, \, \boldsymbol{Q}_l}{\arg \max} \left\langle \boldsymbol{Q}_{l} \mathbf{W}_{l}^{A} \boldsymbol{Q}_{l-1}^{\top}, \, \boldsymbol{P}_{l} \mathbf{W}_{l}^{B} \boldsymbol{P}_{l-1}^{\top}\right\rangle_F +\left\langle \boldsymbol{Q}_{l+1} \mathbf{W}_{l+1}^{A} \boldsymbol{Q}_{l}^{\top}, \,  \boldsymbol{P}_{l+1} \mathbf{W}_{l+1}^{B} \boldsymbol{P}_{l}^{\top}\right\rangle_F \\
    = & \, \underset{\boldsymbol{P}_l, \, \boldsymbol{Q}_l}{\arg \max} \,  \operatorname{tr}\left[\boldsymbol{Q}_{l} \left( \mathbf{W}_{l}^{A} \boldsymbol{Q}_{l-1}^{\top} \boldsymbol{P}_{l-1} \left(\mathbf{W}_{l}^{B}\right)^{\top} + (\mathbf{W}_{l+1}^{A})^{\top} \boldsymbol{Q}_{l+1}^{\top} \boldsymbol{P}_{l+1} \mathbf{W}_{l+1}^{B} \right) \boldsymbol{P}_{l}^{\top}\right]\;.
\end{aligned}
\end{equation}
Similarly, we omit the bias parameter for $\left\langle \boldsymbol{Q}_{l} \mathbf{b}^{A}_{l}, \, \boldsymbol{P}_{l} \mathbf{b}^{B}_{l} \right\rangle_F$, which will be incorporated into the cost summation in the middle of the $\operatorname{tr}$ in the form of $\mathbf{b}^{A}_{l} \left(\mathbf{b}^{B}_{l}\right)^{\top}$.

\end{itemize}

\textbf{Optimization Strategy.} The first case can be reduced to the well-known Linear Assignment Problem (LAP), which can be efficiently solved using the coordinate descent approach~\citep{DBLP:conf/iclr/AinsworthHS23}. For the second case, we reformulate the optimization as in~\autoref{eq:incompatible} and adopt the \textit{alternating optimization strategy} on $\boldsymbol{P}_l$ and $\boldsymbol{Q}_l$: we fix one permutation and optimize the other iteratively, reducing it to a solvable form like the first case.
Since modern neural networks primarily consist of MLPs, our framework can be extended beyond standard dense layers. This includes architectures incorporating residual connections or attention mechanisms for broader applications.

\section{Experiments}

\textbf{Setting of MAL.} We consider five types of model architectures: ones with equal input/output widths, wide-to-narrow, narrow-to-wide, pyramid-shaped, and the inverse of pyramid. Models with 2, 3, 5, and 6 layers are deployed, with widths chosen from \{16, 32, 64, 128\}. The datasets are CIFAR-10, CIFAR-100, MNIST, and FashionMNIST. We employ learning rates of 1$e^{-4}$ and 1$e^{-3}$, selecting the best. In~\autoref{fig:results}, we calculate the area under the loss curve (as a function of lambda) relative to naive merging—lighter colors indicate a more pronounced LMC effect and thus better performance.

\textbf{Bag of Laws for MAL.} \textbf{1}) Merging heterogeneous models on the same dataset is most effective when at least one base architecture performs well. For example, as shown in the first-row of~\autoref{fig:results}, simpler architectures on MNIST avoid overfitting, while on CIFAR-10, a more complex base excels.
\textbf{2}) Across datasets, even models with significantly different architectures can extract knowledge from corresponding layers with varying widths, indicating that zero-padding minimally affects the process since key concepts align correctly.
\textbf{3}) Furthermore, selectively merging only some layers \textit{can preserve LMC} and the semantic progression from shallow to deep layers, whereas merging deep-layer weights into shallower layers may degrade performance (see the right panels of the third and fourth rows).

\section{Conclusion}

In this paper, we introduce Model Assembly Learning (MAL), a novel paradigm that acquires general knowledge by merging heterogeneous parameters from a diverse model zoo layer by layer. Unlike previous methods that require identical architectures, MAL uses a generalized permutation transformation to overcome layer-width mismatches. Our zero-padding and bidirectional alignment strategies preserve original-domain performance within a merging factor threshold while enabling refined knowledge transfer via selective parameter integration. We also cthe larify bag of laws for heterogeneous parameter merging and explore the interplay between different architectures, task characteristics, and LMC. Future work will extend MAL to more complex architectures, multi-task scenarios, and large language models for a more flexible and robust system.

\bibliography{./iclr2025_conference}
\bibliographystyle{iclr2025_conference}

\end{document}